\documentclass[runningheads]{llncs}
\usepackage{cite}
\usepackage{amsmath,amssymb,amsfonts}
\usepackage{algorithmic}
\usepackage{graphicx}
\usepackage{subcaption}
\usepackage{textcomp}
\usepackage{xcolor}

\usepackage{amsmath}
\usepackage{orcidlink}
\usepackage{braket}
\def\BibTeX{{\rm B\kern-.05em{\sc i\kern-.025em b}\kern-.08em
    T\kern-.1667em\lower.7ex\hbox{E}\kern-.125emX}}
\begin{document}

\title{Investigating Parameter-Efficiency of Hybrid QuGANs Based on Geometric Properties of Generated Sea Route Graphs\\
}
\titlerunning{Generating sea route graphs with QuGANs}

\author{
    Tobias Rohe\inst{1}\orcidlink{0009-0003-3283-0586} \and
    Florian Burger \and
    Michael Kölle\inst{1}\orcidlink{0000-0002-8472-9944} \and
    Sebastian Wölckert\inst{1} \and
    Maximilian Zorn\inst{1}\orcidlink{0009-0006-2750-7495} \and
    Claudia Linnhoff-Popien\inst{1}\orcidlink{0000-0001-6284-9286}
}
\authorrunning{Rohe et al.} 

\institute{
    Mobile and Distributed Systems Group, LMU Munich, Germany \\
    \email{tobias.rohe@ifi.lmu.de}
}
\maketitle

\begin{abstract}
The demand for artificially generated data for the development, training and testing of new algorithms is omnipresent. Quantum computing (QC), does offer the hope that its inherent probabilistic functionality can be utilised in this field of generative artificial intelligence. In this study, we use quantum-classical hybrid generative adversarial networks (QuGANs) to artificially generate graphs of shipping routes. We create a training dataset based on real shipping data and investigate to what extent QuGANs are able to learn and reproduce inherent distributions and geometric features of this data. We compare hybrid QuGANs with classical Generative Adversarial Networks (GANs), with a special focus on their parameter efficiency. Our results indicate that QuGANs are indeed able to quickly learn and represent underlying geometric properties and distributions, although they seem to have difficulties in introducing variance into the sampled data. Compared to classical GANs of greater size, measured in the number of parameters used, some QuGANs show similar result quality. Our reference to concrete use cases, such as the generation of shipping data, provides an illustrative example and demonstrate the potential and diversity in which QC can be used.
\end{abstract}


\section{Introduction} \label{introduction}
QC with its probabilistic character and generative AI are two topics of our time that go together surprisingly well. In the event that Moore's Law comes to an end, QC could be a real alternative - assuming improved QC hardware - as many methods of classical generative AI can be transferred to the field of QC. \\
In this study, we look at the successful concept of GANs and how these can be implemented in hybrid form on both classical and QC hardware. Although literature on classical GANs is already extensive \cite{goodfellow2014generative, wang2017generative, creswell2018generative, goodfellow2020generative}, and their use cases are diverse, examples are the generation of brain images \cite{islam2020gan}, the enhancement of image quality \cite{chen2018efficient}, or the estimation of road-traffic \cite{xu2020ge}, there is a lack of literature that looks at the functionalities and abilities of its quantum-classical hybrid version, so-called QuGANs. Like classical GANs, QuGANs consist of a discriminator and a generator, but at least one of the two components is implemented using QC. \\
In our study, we implement the generator of the GAN on a QC-simulator, which then subsequently interacts with a classically implemented discriminator. We learn and generate artificial shipping routes that are geometrically subject to the triangle inequality, which must never be broken. Our primary focus is on the question of how well and how efficiently such hybrid QuGANs can recognise the underlying, simplest geometric structures and reproduce them in the generated data, while approximating with the sampled port distances a bimodal distribution. To learn and represent the underlying geometric structures in the generated data is thereby an ability which is crucial for the successfully execution of many generative tasks \cite{de2018molgan}. Artificial generated data in general can be used as training and / or test instances to develop new tools and algorithms~\cite{islam2020gan}. Here, for example, a future use of our artificially created realistic shipping routes as problem instances for the travelling salesman problem is conceivable. We evaluate the efficiency of our QuGAN implementations with that of a classical GAN with similar, even slightly higher, number of parameters, while we also benchmark the results against a random baseline. This research should help to shed more light on the inherent functioning and capabilities of hybrid QuGANs - motivating further research into this area. \\
Our work is structured as follows: In Sec.~\ref{background} we provide theoretical background for the work at hand. In Sec.~\ref{related_work} we present related work, followed by Sec.~\ref{methodology} and the methodology of our paper. Sec.~\ref{results} presents the results of this work. The paper ends with Sec.~\ref{conclusion} Conclusion.

\section{Background} \label{background}
\subsection{Generative Adversarial Networks (GANs)}
GANs represent a framework in the field of machine learning where two neural network models, the generator ($G$) and the discriminator ($D$), respectively parameterised by sets of neural network weights $\theta_G$, $\theta_D$, are trained adversarially in a zero-sum game. $G_{\theta_G}(z)=x_{fake}$ aims to synthesise fake data $x_{fake}$ from the latent space $z$, that is indistinguishable from a set of real training data $x_{real}$ which the model is trying to replicate. The purpose of training $G$ is thus to newly generate samples that the discriminator mistakes for increasingly realistic data. $D_{\theta_D}(x)$ evaluates the authenticity of both the received real data from the training set and the fake data generated by $G$ with the aim of accurately labeling samples as ``real'' (in the original data set) or not. 
The training process involves $G$ and $D$ simultaneously adjusting their parameters $\theta_G$ and $\theta_D$. The loss of the discriminator $D$ that is incurred in judging the data generated by $G$ is generally quantified via the binary cross-entropy (BCE) loss function: 
\begin{equation}
    \textsc{BCE}(y, d) = -(y * log(d) + (1-y) * log(1-d))
\end{equation}
where $d \in [0;1]$ represents the probability estimated by the discriminator that the sample is real, while $y$ is the associated ground-truth label (1, real or 0, fake). \\ Training continues until a Nash equilibrium is reached where neither $G$ nor $D$ can trivially improve their strategies, meaning that $D$ is not able to discern between real and generated data anymore. The general dynamic of the mini-max game can be described by the following value function:
\begin{align}
\min_G \max_D V(D, G) = & \; \mathbb{E}_{x \sim p_{\text{data}}(x)}[\log D(x)] \nonumber \\
                    & + \mathbb{E}_{z \sim p_z(z)}[\log (1 - D(G(z)))]
\end{align}
where $z \sim \mathcal{N}(0,1)$ is randomly sampled noisy input and $x = G(z)$ is the latent feature vector generated by $G$.\\
One of the main issues with the GAN architecture is mode collapse, where $G$ learns to produce a limited variety of outputs, thus not capturing the full variety of input data. Here, the training process of GANs is unstable, which is characterised by oscillations and non-convergence during the training process. This instability is due to the balance required between $G$ and $D$, where disproportionate gradient updates, produced by e.g. imbalanced learning rates or model sizes,cause one to overpower the other and undermine the training process.

\subsection{Quantum Generative Adversarial Networks (QuGANs)}
In a QuGAN, either $D$ and / or $G$ can be replaced with a variational quantum circuit (VQC) \cite{sim2019expressibility}. In this setup, classical data is mapped to a quantum state using a feature map $f:\mathbb R^m\to\mathcal H^{\otimes n}$, allowing for the application of parameterised quantum gates to generate a desired output state $\ket{\Psi}$. In the end, we obtain the classical data by measuring the quantum system, for instance, in the computational basis. Similar to classical GANs, the parameters are trained against the BCE loss and the minimising direction is calculated via gradient based methods.

\section{Related Work} \label{related_work}
In case of the realisation of only one component of a GAN by means of QC, it is referred to as a hybrid QuGAN \cite{ngo2023survey}. Already early work has speculated that QuGANs will have more diverse representation power than their classical counterparts, particularly for very high-dimensional data \cite{dallaire2018quantum, lloyd2018quantum}. Over time, various realisations of QuGANs such as Tensor-Network-Based GANs \cite{huggins2019towards}, Quantum Conditional GANs \cite{dallaire2018quantum} and Quantum Wasserstein GANs \cite{chakrabarti2019quantum} have been developed. The application of QuGANs can be diverse, from approximating quantum states \cite{chakrabarti2019quantum}, to generating discrete distributions \cite{situ2020quantum}, to learn and load random distributions \cite{zoufal2019quantum}, to generating images \cite{stein2021qugan}. \\
Comprehensive studies have already been carried out on the efficiency, especially the parameter efficiency, of QuGANs. These studies have investigated how QuGANs perform in comparison to classic GANs with the same number of parameters. Kao et al. (2023) \cite{kao2023exploring} and Stein et al. (2021) \cite{stein2021qugan} demonstrate that QuGANs achieve comparable or superior performance with substantially fewer parameters. Kao et al. highlight ongoing challenges in generating unique and valid molecules. Li et al. (2021) \cite{li2021quantum} further emphasise the efficiency of QuGANs, showing that their models can learn molecular distributions effectively with a reduced parameter count and improved training processes through the use of multiple quantum sub-circuits. Additionally, Anand et al. (2021) \cite{anand2021noise} explore noise resistance in hybrid quantum-classical setups, finding that QuGANs can maintain performance despite moderate noise levels, which is pivotal for practical applications. \\
However, to the best of our knowledge, there is no research into the creation of graphs using QuGANs, although some models do have connections to graphs, especially the underlying graph structures of their application, such as for generative chemistry \cite{de2018molgan}. In the field of classical GANs, on the other hand, there is already literature and various adapted models. A survey by Zhu et al. (2022) \cite{zhu2022survey} describes the current state of research on graph generation via deep learning models, also looking at how classical GANs can be used for this purpose.

\section{Methodology} \label{methodology}
Our methodology builds on a synthesis of concepts from two foundational pipelines: a pipeline for quantum machine learning~\cite{gabor2020holy} and a pipeline for quantum optimization~\cite{rohe2024problem}, ensuring a comprehensive approach to the problem.

\subsection{Data Preprocessing}
We employed the Python package \texttt{searoute} to calculate realistic sea-routes between different ports \cite{searoute}. For the underlying training dataset, we randomly sampled four ports from the $3,669$ ports of the \texttt{searoute} package and computed the distances between all pairs sampled, creating a fully connected graph, where each node represents a port, and the edge weights denotes the minimum distances between those ports. To avoid extreme cases where two ports are directly next to each other, all graphs where two ports were closer than $100$ nautical miles to each other were excluded. This follows from the logic that the triangle inequality can be broken particularly easily in those cases. We sampled a total of $1,000$ graphs for our training dataset and pre-processed our data by normalising the sum of edges to one, while preserving the integrity of the distance relationships. \\
\subsection{Classical GAN}
The discriminator used for our classical GAN, but also for the later described QuGANs, consists of a fully connected neural network with three linear layers: an input layer, a hidden layer with $16$ neurons, and an output layer. The network processes tensors representing graph data, with an input size of $6$, corresponding to the edges in our fully connected four-port graphs. Each layer, except for the output layer, is activated by a LeakyReLU function to mitigate the issue of dying gradients \cite{maas2013rectifier}. The output layer utilises a sigmoid activation function to classify the inputs as real or fake. The discriminator has a total of $129$ trainable parameters.
The classical generator mirrors this architecture but includes a hidden layer of $10$ neurons and receives a noise vector sampled from a normal distribution as input. The output is a tensor matching the dimensions of real graph data, activated by a ReLU function to ensure non-negative edge weights. The generator has a total of $136$ trainable parameters. \\
\subsection{QuGAN}
For each of the four hyrbid QuGAN models tested, we employed the classical discriminator described above and constructed four different quantum generators. For the quantum generator, the latent variable was sampled from a standard normal distribution and then encoded into a quantum state using angle embedding, leading us to utilise six qubits. We test two different quantum circuits (Ansätze) based on QISKIT's efficient $\mathrm{SU}(2)$ $2$-local circuit, each one with 5 and with $10$ layers, giving us four generator models. The first circuit consists of layers of one ladder rotational-X gates and one ladder Pauli-Y gates, followed by a circular CNOT entanglement. The second circuit is less restrictive and consists of layers of two ladders of rotational-X and rotational-Y gates, and again a circular CNOT entanglement. The Ansätze utilizing the Pauli-Y gates have $36$ and $66$ parameters for the different numbers of layers, while the Ansätze with rotational-Y gates have $72$ and $132$ parameters. We will refer to them as QuGAN followed by their respective number of parameters, e.g. QuGAN($36$). The output of these circuits calculate the probability that each qubit is in the state $\ket{0}$, which represents an edge weight, therefor the distance between two ports. We re-normalised the measurement results to the sum of one, so that the edge-ratios remain the same, but the measurement data do not differ from the real training data by their sum. Our numerical simulations were performed without noise. \\ 
\subsection{Training \& Evaluation}
All models were trained using the Adam optimiser over $1,000$ epochs, with a batch size of $32$. The learning rates were set at $0.3$ for the discriminator and $0.001$ for the generator, optimising against the BCE loss. This training process was repeated across five different seeds, and the average outcomes were reported. \\
To assess the validity of the generated graphs, we applied the triangle inequality test. Valid graphs were those where the post-processed edge weights comply with the triangle inequality.
For each vertice triple $A,B,C \in V$ it hold that $d(A, B) \leq d(A, C) + d(C, B)$, with a distance function $d : \mathbb{R} \times \mathbb{R} \to \mathbb{R}$.

As a further additional baseline, we sampled $1,000$ graphs from the training data's post-processed weight distribution using kernel density estimation, providing a reference for generating random graphs based on the training dataset. This should help us to estimate how easy or hard it is to violate the triangle inequality if the edges are blindly sampled from the distribution without considering their geometric relations to each other.

\section{Results} \label{results}
\begin{figure*}[!t]
  \centering
  \begin{subfigure}[t]{0.49\textwidth}
    \centering
    \includegraphics[width=\linewidth]{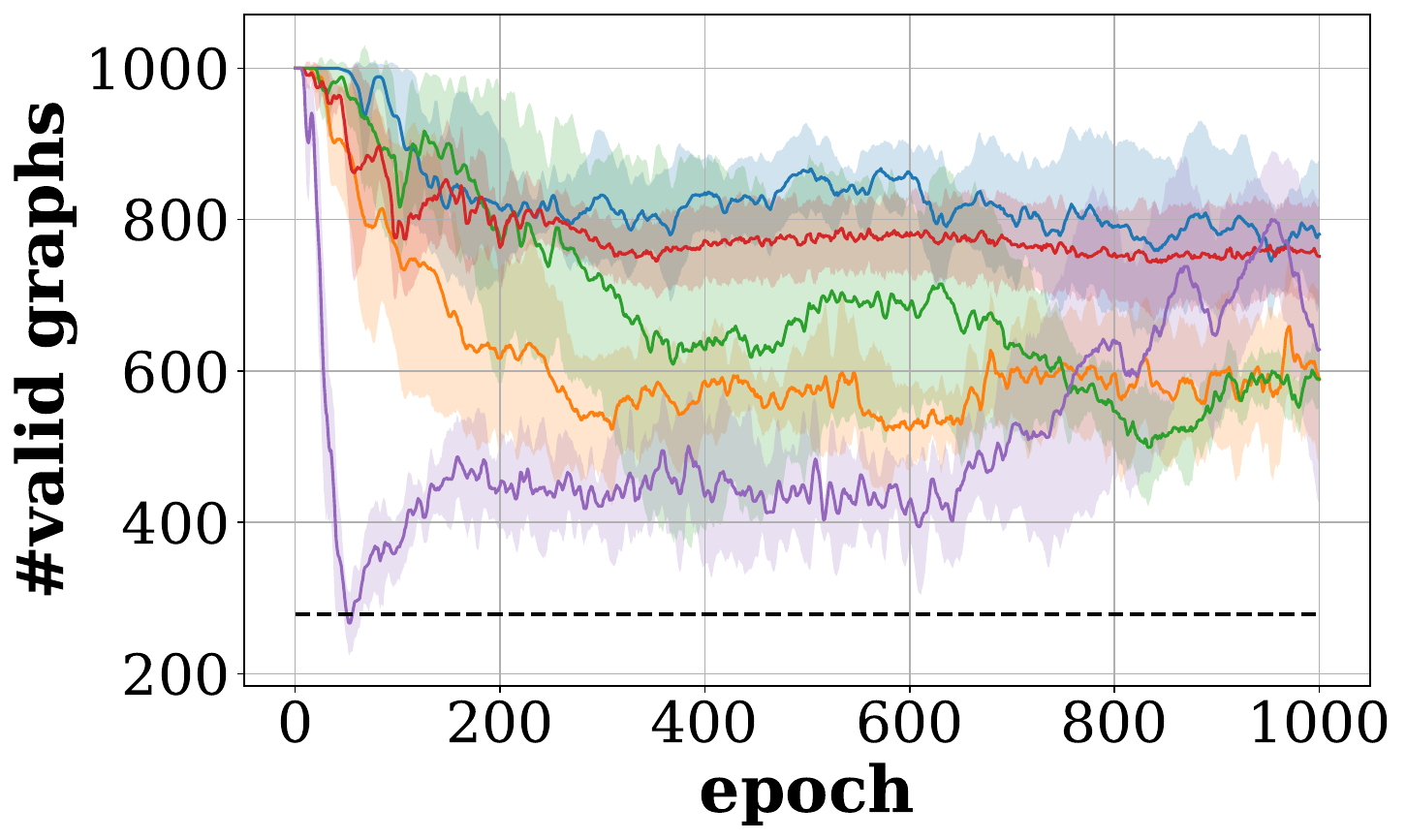}
    \caption{Average Number of Valid Graphs}
    \label{fig:avg_no_valid_graphs_wo}
  \end{subfigure}%
  \hfill
  \begin{subfigure}[t]{0.49\textwidth}
    \centering
    \includegraphics[width=\linewidth]{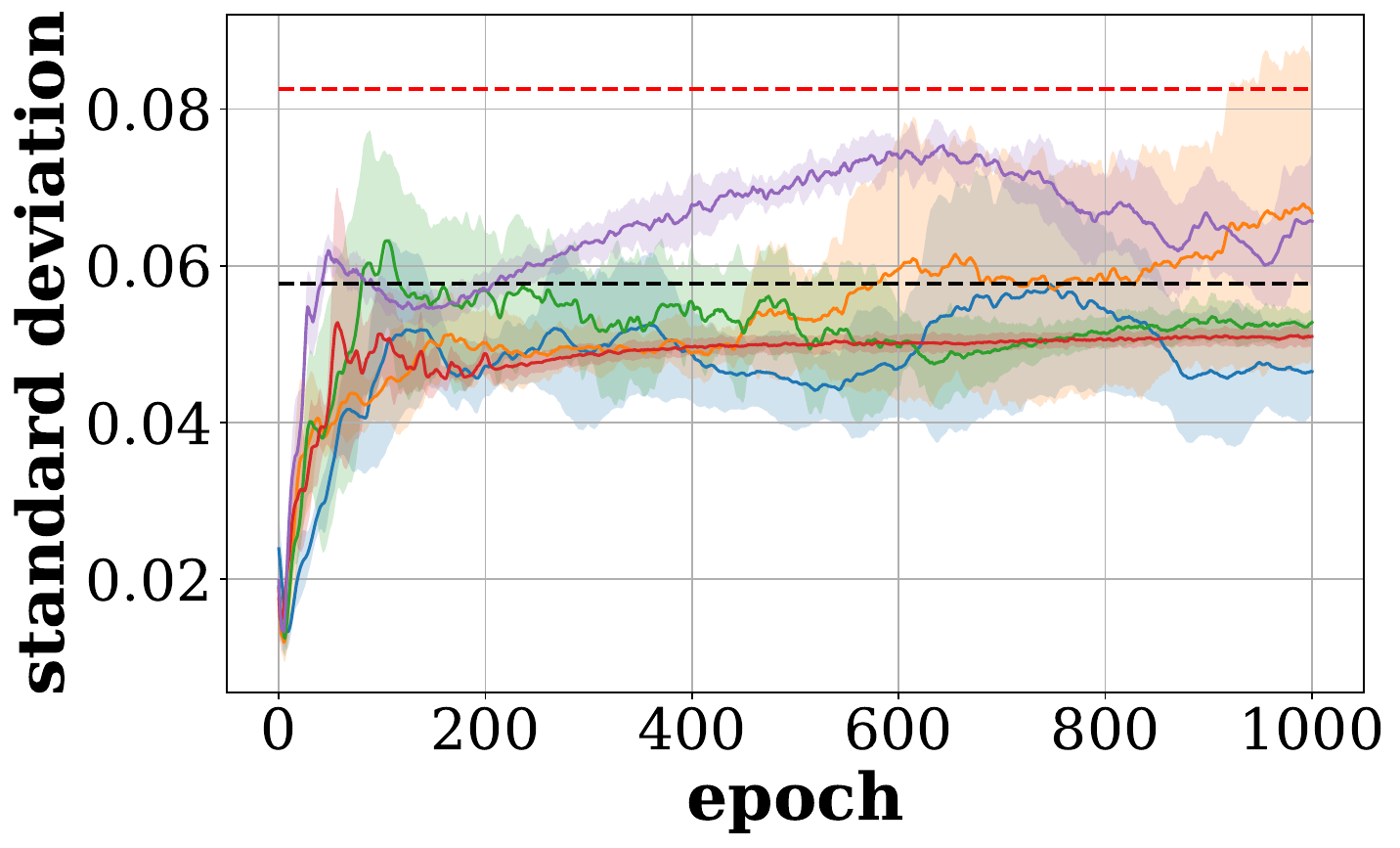}
    \caption{Average Standard Deviation of Edge-Weights}
    \label{fig:avg_stddev_weights_wo}
  \end{subfigure}

  \bigskip
  
  \begin{subfigure}[t]{0.49\textwidth}
    \centering
    \includegraphics[width=\linewidth]{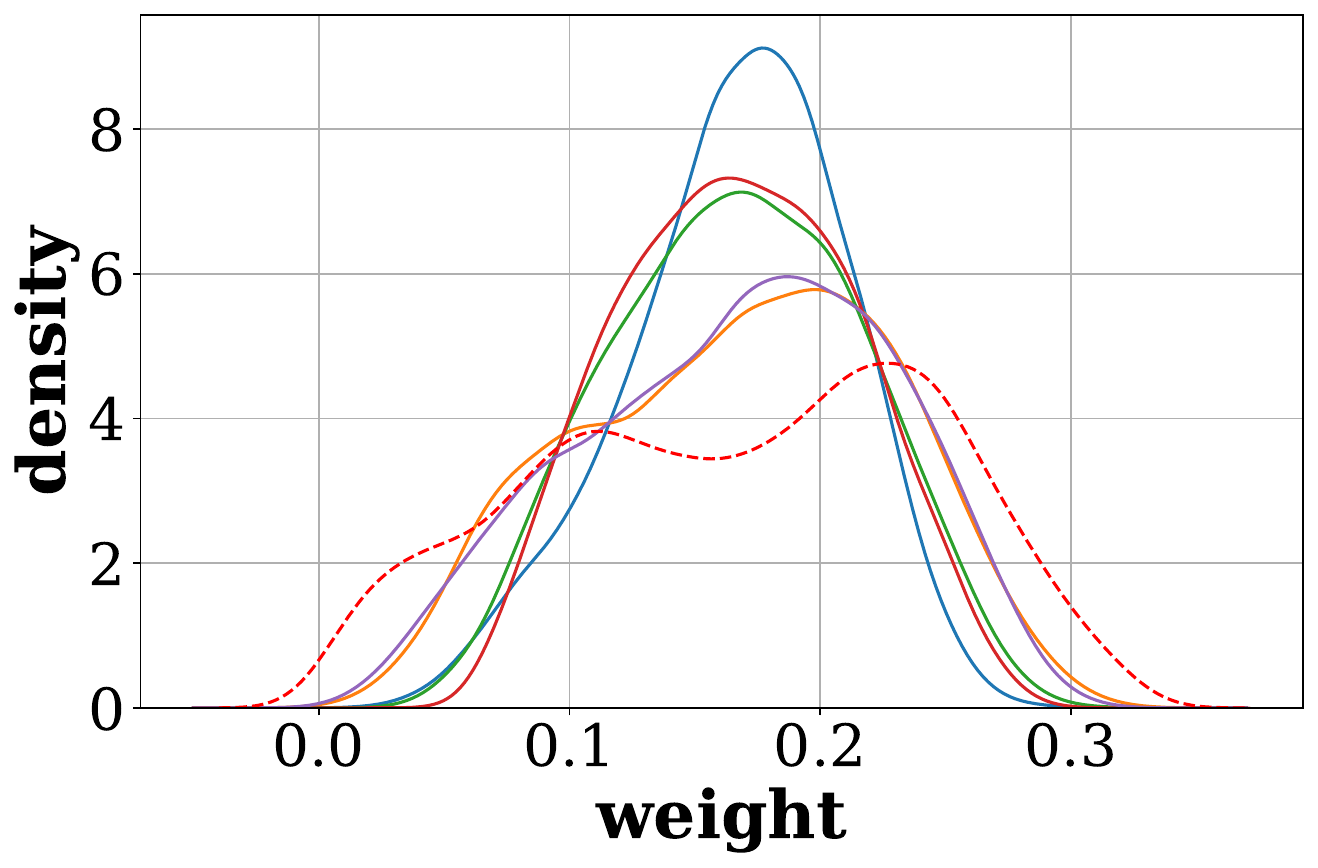}
    \caption{Edge-Weight Distribution}
    \label{fig:data_weight_distri_wo}
  \end{subfigure}%
  \hfill
  \begin{subfigure}[t]{0.49\textwidth}
    \centering
    \includegraphics[width=\linewidth]{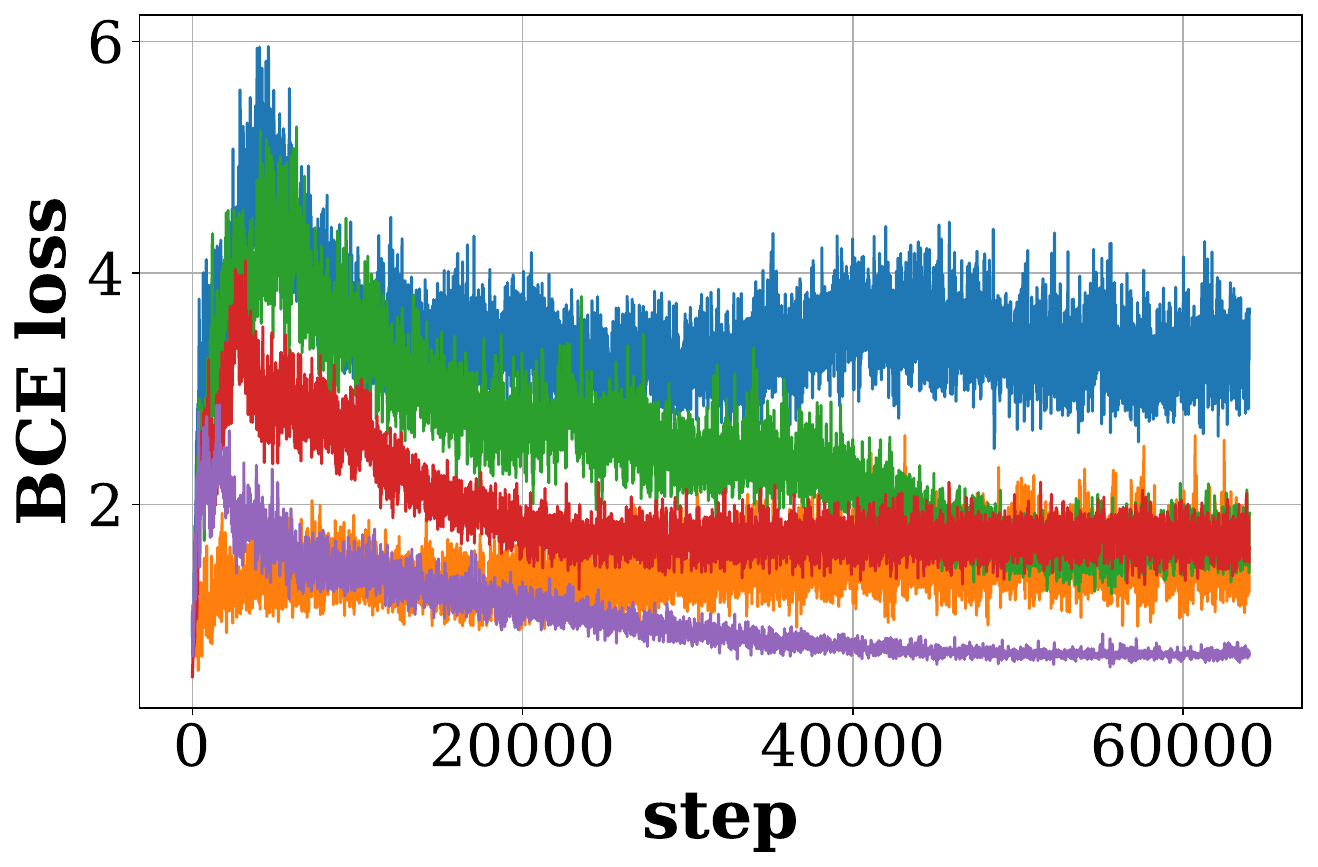}
    \caption{Average Loss of Generator}
    \label{fig:avg_loss_generator_wo}
  \end{subfigure}

  \bigskip

  \begin{subfigure}[t]{\textwidth}
  \centering
  \includegraphics[width=\linewidth]{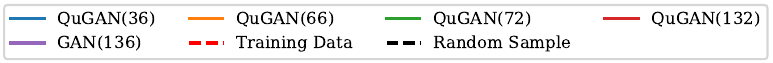}
  \label{fig:legend}
  \end{subfigure}%

  \caption{Subgraph a) shows the development of the number of valid graphs over the epochs. Subgraph b) illustrates the development of the calculated standard deviation of the weights generated over the epochs. Subgraph c) shows the density of the edge weights, while finally, subgraph d) visualises the loss of the generators.}
  \label{fig:results}
\end{figure*}

In the following, we will take a closer look at our results presented in Fig. \ref{fig:results}. We first look at the results in the light of valid graphs (Fig. \ref{fig:avg_no_valid_graphs_wo}), then at the standard deviation (Fig. \ref{fig:avg_stddev_weights_wo}) and the density functions of the sampled edge weights (Fig. \ref{fig:data_weight_distri_wo}), and finally at the losses achieved by the different generators (Fig. \ref{fig:avg_loss_generator_wo}).

At the beginning, all implementations start with approximately 1000 valid graphs, which then relatively fast decline within the first 100 epochs, but then recover again. In general, it can be noted that the QuGANs keep up with the classical implementation in terms of valid graphs as evaluation metric. The QuGAN(36) and QuGAN(132) implementation show very stable and good (approx. 80\% valid graphs) results. The QuGAN(66) and QuGAN(72) implementations are less stable and show significantly less valid graphs (approx. 60\% valid graphs). The classical GAN, on the other hand, takes much longer to reach a solid level and ultimately fluctuates between 60\% and 80\% valid graphs. However, it should be noted that all (Qu)GAN implementations clearly outperform random sampling with its 27.9\% valid graphs baseline.

The standard deviation of the edge weights in the training dataset was 0.0826, which also should be observed in our results. However, all implementations start with a very low standard deviation (close to zero), and increase it over the subsequent training process. These results explain the observation of the development of the valid graphs in Fig. \ref{fig:avg_no_valid_graphs_wo}, where all (Qu)GANs start at around 100\% valid graphs and then drop sharply. At the beginning of the training process, all (Qu)GAN implementations produce very monotonous edge weigths without strong variances and deviate from this behaviour later in the training process. While the generation of similar edge weights ensures that the triangle inequality is not broken, it makes it for the discriminator a simple task to distinguish between real and fake data. With the increase of variance of edge weights, the differentiation becomes more difficult, but the triangle inequality is also increasingly broken. Our results indicate, that the QuGANs in general seem to have a harder time in terms of generating variance, with only implementation QuGAN(66) showing a similar, even higher standard deviation compared to the classical GAN towards the end of training. Implementations QuGAN(36), QuGAN(132) and QuGAN(72) are clearly behind here. 

Looking at the edge-weight distribution, the classical GAN and the hybrid QuGAN(66) implementation approximate the bimodal nature of the underlying training data (dashed-red line) more effectively than other models, although they do not sharply replicate the two peaks. These models show a more pronounced but smoother rise towards the second peak, with a centre slightly offset from the training data. Although these implementations do not match the exact height of the training data peaks, as well as the variability, they come closest to capturing the overall trend, despite the evident shift and reduced peak magnitude in the first peak. Conversely, the models represented by the QuGAN(132) and QuGAN(72) curves depict a unimodal distribution with a peak situated between the two actual peaks of the training data, suggesting a simpler but less accurate modeling of the dataset. The QuGAN(36) model also follows this trend but with a more pronounced central peak, indicating a tighter concentration around the median weigh.

In all our implemented (Qu)GAN versions, we use the same discriminator, which is why we attribute a certain significance to the comparison of the average losses of the generator compared to the other implementations. The classical GAN performs best here. This can be interpreted with caution as meaning that the classical generator is best able to build artificial graphs that cannot be distinguished by the discriminator, although the feedback loop of the training effect of D by G must also be mentioned here. The best quantum implementation is QuGAN(66), closely followed by QuGAN(72) and QuGAN(132). The QuGAN(36) implementation performs the worst, although we had the best results here in terms of evaluating the triangle inequality.


\section{Conclusion} \label{conclusion}
In our, to our knowledge, unique problem setting, we examined QuGANs for their parameter efficiency. In particular, we chose a setup where not only the modelling of a density function was in the foreground, but in each instance several values were sampled from the distribution, which were all directly connected and dependent on each other via the triangle inequality. With the setting applied, not only the sampling, but also the underlying geometric properties are included in the evaluation and comparison. The best-performing quantum generators were QuGAN(66) and QuGAN(132), which showed a clear difference in the learned weight distribution. Both Ansätze are built using the same generators, but QuGAN(66) has fixed Pauli-Y gates. This reduced expressibility allows for easier training while still being capable of approximating the bimodal distribution of the real data. Overall, our results suggest that QuGANs struggle to reflect variance in their generated data. The normalisation of the generated graphs plays an important role in this context. While it reduces the variance of the generated weights, it also decreases the variance of the training data. Although GANs are generally difficult to train, this approach enabled a more stable training process. At the same time, they can still generate balanced results in the area of tension between the variation in the edge weights and the fulfilment of the triangle inequality. For example, the QuGAN(66) implementation shows equally good results in terms of variance and valid graphs compared to the classical GAN, although less than half as many parameters were used. This is in line with previous work on QuGANs and parameter efficiency, although our work extends this with underlying geometric properties. In particular, the underlying task chosen here has great potential for future research. The search for circuits and implementations that can better reflect the variance of the training data as well as their geometric properties should be the main focus of future research. New circuit designs~\cite{stein2023introducing}, training approaches~\cite{kolle2024study}, but also new baselines with fewer or no entanglements~\cite{rohe2024questionable} can be of interest here.

\section*{Acknowledgement}
This paper was partially funded by the German Federal Ministry of Education and Research through the funding program "quantum technologies - from basic research to market" (contract number: 13N16196). Furthermore, this paper was also partially funded by the German Federal Ministry for Economic Affairs and Climate Action through the funding program "Quantum Computing -- Applications for the industry" (contract number: 01MQ22008A).


\vspace{12pt}

\end{document}